\documentclass{article}
\usepackage{spconf,amsmath,graphicx,hyperref}
\usepackage{times}
\usepackage{latexsym}
\usepackage{booktabs,multicol,multirow}
\usepackage{amsfonts}
\usepackage{amsmath}
\usepackage{booktabs}
\usepackage{xcolor}
\usepackage{colortbl}
\usepackage{setspace}

\title{LLMPopcorn: Exploring LLMs as Assistants for Popular Micro-video Generation}
\name{
    \begin{tabular}{c}
        Junchen Fu$^{1}$, Xuri Ge$^{2*}$\thanks{* Corresponding author}, Kaiwen Zheng$^{1}$, Alexandros Karatzoglou$^{3}$, Ioannis Arapakis$^{4}$, \\
        Xin Xin$^{2}$, Yongxin Ni$^{5}$, and Joemon M. Jose$^{1}$
    \end{tabular}
}
\address{
    $^{1}$University of Glasgow, United Kingdom \\
    $^{2}$Shandong University, China \\
    $^{3}$Amazon, Spain \\
    $^{4}$Telefónica Research, Spain\\
    $^{5}$National University of Singapore, Singapore\\
}

\begin{document}
\ninept
%
\maketitle
%
\begin{abstract}

In an era where micro-videos dominate platforms like TikTok and YouTube, AI-generated content is nearing cinematic quality. The next frontier is using large language models (LLMs) to autonomously create viral micro-videos, a largely untapped potential that could shape the future of AI-driven content creation. To address this gap, this paper presents the first exploration of  \underline{LLM}-assisted \underline{pop}ular mi\underline{c}ro-vide\underline{o} gene\underline{r}atio\underline{n} (LLMPopcorn\footnote{We selected popcorn as the icon for this paper because it symbolizes leisure and entertainment. This aligns with this study on leveraging LLMs as assistants for generating popular micro-videos, which are often consumed during leisure time.}). Specifically, we empirically study the following research questions: \textit{(i) 
How can LLMs be effectively utilized to assist popular micro-video generation?
(ii) To what extent can prompt-based enhancements optimize the LLM-generated content for higher popularity?
(iii) How well do various LLMs and video generators perform in the popular micro-video generation task?
} Exploring these questions, we show that advanced LLMs like DeepSeek-V3 can generate micro-videos with popularity rivaling human content. Prompt enhancement further boosts results, while benchmarking highlights DeepSeek-V3 and R1 for LLMs, and LTX-Video and HunyuanVideo for video generation. This work advances AI-assisted micro-video creation and opens new research directions. The code is publicly available at \url{https://github.com/GAIR-Lab/LLMPopcorn}.
\end{abstract}
\begin{keywords}
Micro-video Generation, Video Popularity, LLM, LLMPopcorn
\end{keywords}
\section{Introduction}
Micro-videos  (or short videos)  have emerged as a crucial element of the digital economy, representing a multi-billion-dollar industry \cite{guo2024survey}. They have become an integral part of daily life for people worldwide, providing substantial commercial value for social media platforms and content creators. Popular content creators can receive significant revenue through their content \cite{ran2022revenue}, which underscores the ever-growing influence of micro-videos in modern society.

Despite their widespread popularity and financial impact, producing popular micro-video content remains a costly and labor-intensive process. Professional filming, scripting, and editing require significant time and resources, which not all creators or businesses can afford.
Driven by these challenges, we explore simplifying the micro-video creation process using the state-of-the-art AI solutions, with a particular focus on generating popular micro-videos.

On one hand, the rapid advancement of Large Language Models (LLMs) like ChatGPT has unlocked new possibilities for content generation, demonstrating strong capabilities in tasks such as document summarization \cite{jin2024comprehensive}, programming \cite{cai2023low}, and recommendation~\cite{fu2024iisan,fu2024exploring,fu2025crossan,yuan2023go}. On the other hand, video generation models powered by diffusion techniques and neural rendering are transforming creative content production, enabling high-quality synthesis for applications in film production and interactive media \cite{liu2024sora}.
Significant breakthroughs in LLMs and video generation models have prompted many studies \cite{zhou2024survey,han2025video} to integrate LLMs to facilitate the capability of automatic video generation. 
For instance, VideoLLMs like VideoPoet~\cite{kondratyuk2023videopoet} and GPT4Video~\cite{wang2024gpt4video} leverage LLMs and video generation models for multimodal video understanding and generation. 

Current studies on LLMs for video generation overlook their role in creating popular micro-videos, focusing instead on quality metrics like resolution and duration. Recent research, including models like those in \cite{lu2024mufm} and datasets like Microlens \cite{ni2023content}, emphasizes micro-video popularity prediction. However, the use of mainstream LLMs to generate highly popular micro-videos, leveraging their text generation and planning capabilities \cite{wu2025automated,fu2025efficient}, remains underexplored.

Driven by the above insights, we present the first exploration of the research question of \textit{whether LLMs can assist in the \textbf{popular} micro-video creation.} To address this research question systematically, we propose and investigate the following three Research Questions (RQs):

\noindent\textbf{RQ1:} \textit{How can LLMs be effectively utilized to assist popular micro-video generation?} \\
We present LLMPopcorn, a micro-video creation pipeline that uses LLMs to generate prompts for popular video production. To assess popularity, we integrate a state-of-the-art offline evaluator and compare three mainstream LLMs against human-created videos.

\vspace{0.2cm}

\noindent\textbf{RQ2:} \textit{To what extent can prompt-based enhancements optimize the content generated by LLMs for achieving greater popularity?} \\
We propose a \textbf{Prompt Enhancement} combining Retrieval-Augmented Generation (RAG)~\cite{lewis2020retrieval} and chain-of-thought (CoT) prompting~\cite{wei2022chain}, inspired by human content creation.

\vspace{0.2cm}

\noindent\textbf{RQ3:} \textit{How do various LLMs and video generators perform on popular micro-video generation?} \\
We compare five LLMs and three video generation models using the LLMPopcorn pipeline.

By investigating these research questions, this study represents an initial exploration of this emerging domain, providing key insights into AI-assisted micro-video creation.

\section{The Proposed \textit{LLMPopcorn}}

\begin{figure}[t]
    \centering
    \includegraphics[width=\linewidth]{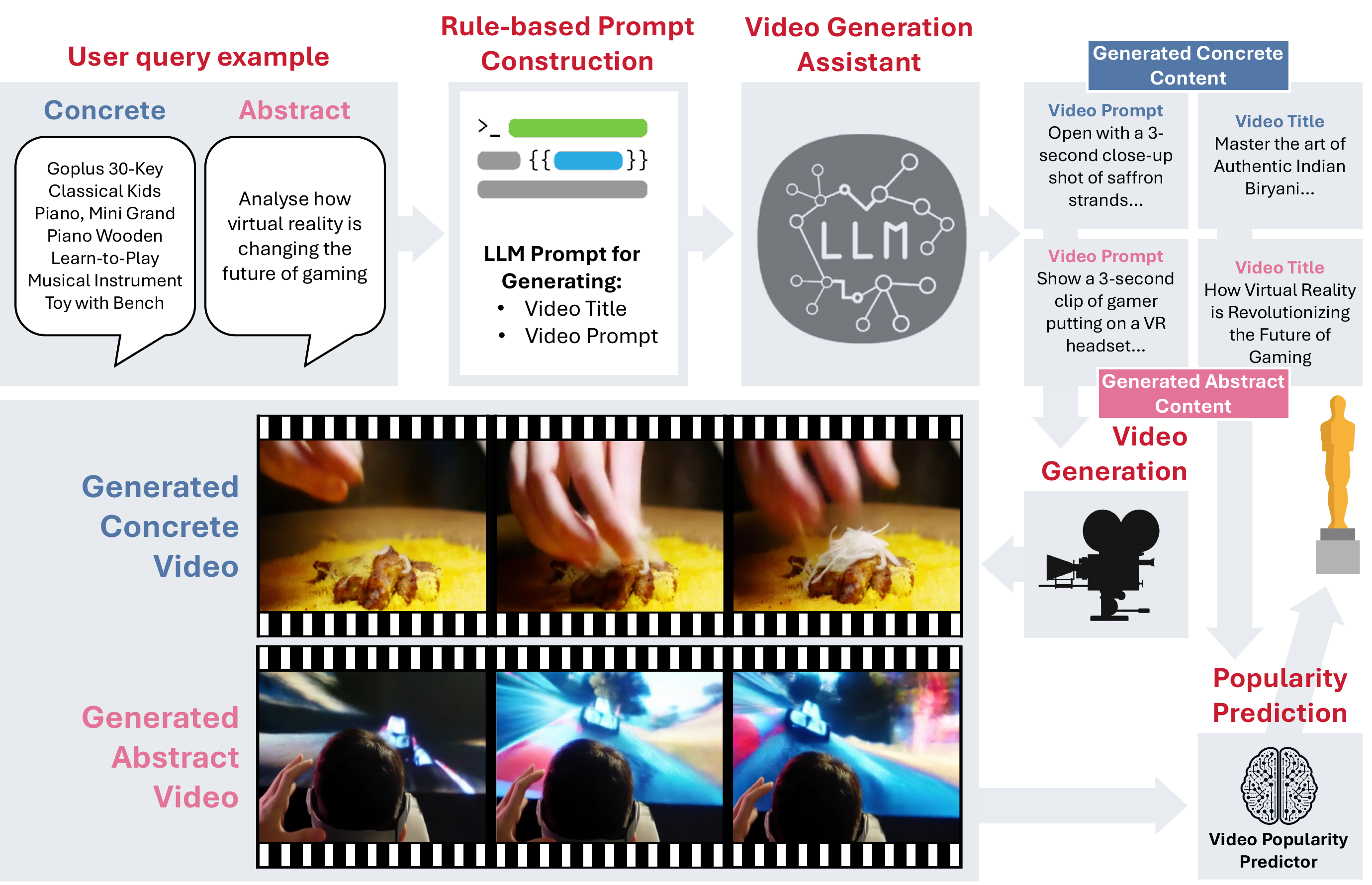}
    \caption{Overview of the LLMPopcorn pipeline. An LLM processes a user query, generating a video title and prompt for a video generator to create a micro-video. The video and title are then evaluated for predicted popularity.}
    \vspace{-0.1in}
    \label{fig:pipeline}
\end{figure}

\subsection{Problem Formulation}

We consider a problem in which a user provides an input prompt \(x \in \mathcal{X}\). An LLM processes this prompt and generates both a video title and a video prompt, which we denote by \((t, p) = f(x)\), where \(t\) belongs to the set of titles \(\mathcal{T}\) and \(p\) belongs to the set of video prompts \(\mathcal{P}\). Next, a pre-trained video generation model \(g\) uses the generated prompt to produce an AI-created micro-video \(v \in \mathcal{V}\), that is, \(v = g(p)\). An offline video popularity predictor \(h\) then assigns a popularity score \(s \in \mathbb{R}\) to the generated video, so that \(s = h(v,t)\). Overall, the entire pipeline can be represented as the composite function: 
\(
F(x) = h\bigl(g(f(x))\bigr).
\)

The objective is to optimize the LLM function \(f\) (with \(g\) and \(h\) fixed) to maximize the expected popularity score of the generated videos. Formally, we wish to achieve:
\(
\max_{f \in \mathcal{F}} \; \mathbb{E}_{x \sim \mathcal{X}} \left[ h\bigl(g(f(x))\bigr) \right],
\)
where \(\mathcal{F}\) denotes the set of all possible configurations of the LLM. This formulation captures the aim of leveraging LLMs to enhance the creation of popular AI-generated micro-videos.

\subsection{Pipeline}
 Based on the problem formulation, we propose a new \textbf{LLMPopcorn} pipeline in Figure \ref{fig:pipeline} 
 that begins by constructing user prompts, categorized into concrete and abstract prompts, which are constructed by rules into inputs for an LLM. The LLM serves as a video generation assistant, producing \textit{video titles} and \textit{video generation prompts} based on these inputs. These video generation prompts are then fed into a video generator to create corresponding videos. The generated videos are combined with their titles to form final short videos, which are subsequently evaluated using a pre-trained video popularity prediction model to assess their potential impact. This pipeline streamlines the process of AI-driven video generation and evaluation, enabling automatic popular micro-video content creation.

\begin{figure}[t]
    \centering
    \includegraphics[width=\linewidth]{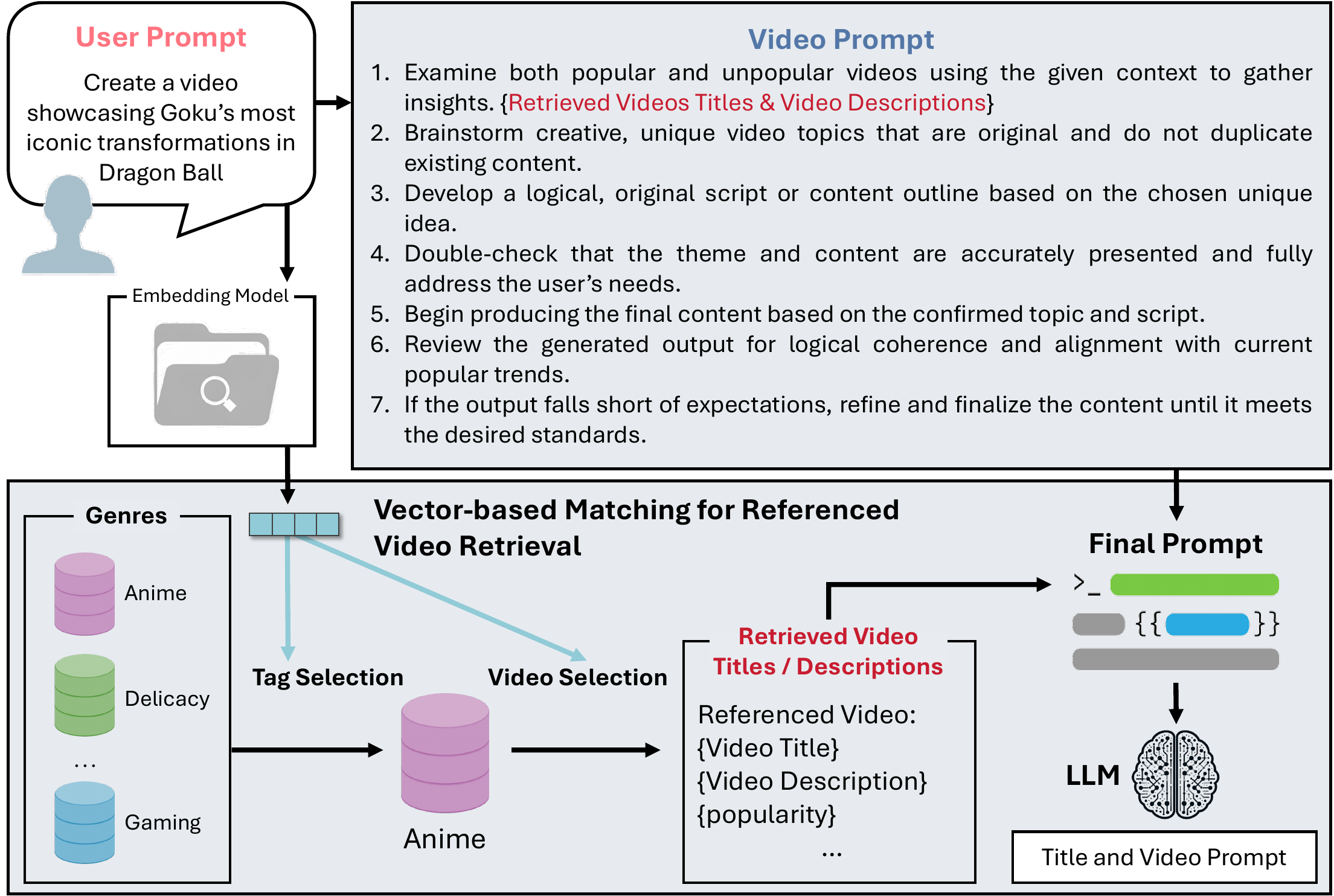}
    \caption{An overview of the Prompt Enhancement (PE) process. PE enables the LLM reviewing relevant micro-videos from the database and engaging in chain-of-thought reasoning.}
    \label{fig:PE_process}
\end{figure}

\subsection{Prompt Enhancement}

We demonstrate the Prompt Enhancement process in Figure \ref{fig:PE_process}, which enhances the input prompt for LLM inspired by human content creators. Formally, let \(x \in \mathcal{X}\) be the user prompt (e.g., ``Create a video showcasing Goku’s  most iconic transformations in Dragon Ball."). An embedding function \(f_{\text{embed}}: \mathcal{X} \to \mathbb{R}^d\) maps \(x\) to its vector representation:
\(
\mathbf{x} = f_{\text{embed}}(x).
\)
Given a predefined set of topic tags \(\mathcal{T}\) (e.g., Food, Anime, Movie) with corresponding embeddings, a matching algorithm based on cosine similarity is used to select the most relevant tags:
\(
\mathcal{T}_x = \operatorname{arg\,max}_{t \in \mathcal{T}} sim\bigl(\mathbf{x}, f_{\text{embed}}(t)\bigr),
\)
where \(sim(\cdot,\cdot)\) denotes the cosine similarity. For each selected tag \(t \in \mathcal{T}_x\), the system retrieves Top-K associated video samples from the test set of the Microlens dataset\footnote{Since MMRA is trained on the training set of the Microlens dataset, we exclusively use the test set to prevent data leakage.} and partitions them into positive examples \(S^+_t\) and negative examples \(S^-_t\) based on popularity. Finally, an enhancement function constructs the new prompt:
\(
x' = \operatorname{Enhance}\Bigl(x,\ \bigcup_{t \in \mathcal{T}_x} \{S^+_t,\, S^-_t\}\Bigr),
\)
which is then used to guide the video generation.

\section{Experiment Setup}\label{sec:dataset}

\noindent\textbf{Dataset.} To enhance comprehensiveness, we designed two types of user prompts as inputs for the LLM: \textbf{Concrete} and \textbf{Abstract}. The user prompts are primarily derived from five key categories—Anime, Delicacy, Daily Sharing, Film \& Television, and Gaming. We reference the Microlens dataset \cite{ni2023content}, as it is the largest available micro-video dataset. Concrete prompts include specific elements that provide clear details, while abstract prompts describe videos in a more general and high-level manner. Specifically, we utilized ChatGPT-4o~\cite{hurst2024gpt} to generate two corresponding user prompt datasets, each containing \textit{one hundred} prompts, based on instructions to include either concrete elements or general content.\footnote{Due to the high cost of video generation, overall two hundred prompts are similar size to existing video generation datasets.\cite{chivileva2024dataset}}
These datasets are further divided into the aforementioned five categories, with each category comprising twenty prompts. This approach ensures a balanced and diverse set of prompts across all categories, catering to both specific and broad input styles for the LLM. 

\noindent\textbf{Evaluation.}
In this study, we assess the popularity of generated micro-videos using the MMRA model \cite{zhong2024predicting}, defining popularity as the median number of comments for robustness, as per \cite{zhong2024predicting}. Higher values indicate greater median engagement, suggesting videos with higher engagement potential. In model-wise comparisons (e.g., Section \ref{sec:PE} and Section \ref{sec:Bench}), we use Win-Rate \cite{gao2024bayesian} to evaluate model performance in predicting popularity, comparing micro-videos generated by Model A and Model B for a given user prompt. This pairwise approach is common in the literature \cite{gao2024bayesian}. Additionally, we conducted a user study following \cite{cheng2024emotion}, involving five expert volunteers with master’s degrees and over five years of experience in micro-video consumption and creation.

\begin{table}[t]
    \centering
    \renewcommand{\arraystretch}{1}
    \setlength{\tabcolsep}{2pt}
    \renewcommand{\arraystretch}{0.7}
    \caption{Comparison of popularity across different categories and LLMs.}
    \begin{tabular}{l|ccc|ccc}
        \toprule
        \multirow{2}{*}{Category} & \multicolumn{3}{c|}{Abstract} & \multicolumn{3}{c}{Concrete} \\
        \cline{2-7}
        & Llama & DeepSeek & GPT & Llama & DeepSeek & GPT \\
        \midrule
        Anime                        & 0.54  & \textbf{0.92}  & 0.63  & 0.37  & \textbf{0.45}  & 0.39  \\
        Daily Sharing                & \textbf{0.59 } & 0.36  & 0.23  & 0.28  & \textbf{0.69}  & 0.61  \\
        Delicacy                     & 0.48  & \textbf{1.12}  & 0.64  & \textbf{0.66}  & 0.59  & 1.24  \\
        Film    & 0.68  &\textbf{1.02}  & 0.82  & 0.46  & \textbf{0.66}  & 0.35  \\
        Game                         & 0.38  & 0.49  & \textbf{0.55}  & 0.35  & 0.30  & \textbf{0.56}  \\
        \bottomrule
    \end{tabular}
         \vspace{-0.2in}
    \label{tab:comparison_categories}
\end{table}

\begin{table}[t]
\centering
\renewcommand{\arraystretch}{0.7}
\caption{Comparison of popularity medians for video generation tasks (using CogVideoX-5b as the video generator). Human-created videos are included for reference. The popularity predictor provides predicted values of Human Crafted video for comparison. Concrete-M and Abstract-M represent the human videos from the Microlens dataset that are most closely matched to the same user prompt in the concrete and abstract datasets, respectively.}
\begin{tabular}{@{}lcc@{}}
\toprule
\textbf{LLMPopcorn}            & \textbf{Concrete} & \textbf{Abstract} \\ \midrule
\textbf{Llama-3.3-70B}  & 0.41              & 0.57              \\
\textbf{DeepSeek-V3}    & \textbf{0.56}              & \textbf{0.73}              \\
\textbf{ChatGPT-4o}     & 0.52              & 0.49      \\
\midrule

\midrule
\textbf{Human Video} & \textbf{Concrete-M} & \textbf{Abstract-M}\\
\midrule
\textbf{Microlens}     & 0.44              & 0.71     
\\ \bottomrule
\end{tabular}
\label{tab:popularity_medians}
\vspace{-0.1in}
\end{table}

\begin{figure}[t]
    \centering
    \includegraphics[width=\linewidth]{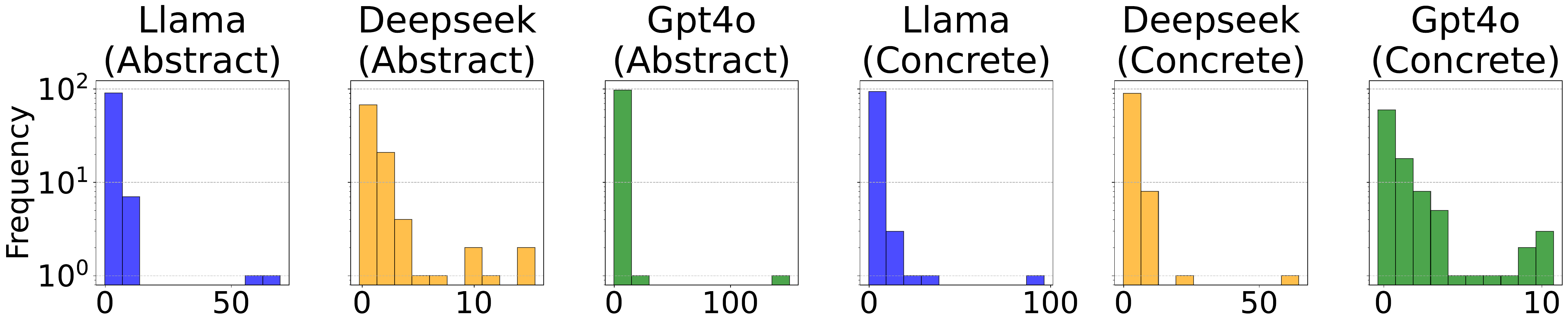}
    \caption{Predicted popularity distributions for different LLMs. Y-axis: frequency. X-axis: predicted popularity.}
    \label{fig:popularity_distribution}
\end{figure}
\begin{table}[t]
\centering
\renewcommand{\arraystretch}{0.65}
\caption{Comparison of popularity for Prompt Enhancement's win rates across different LLMs.}
\begin{tabular}{@{}llc@{}}
\toprule
\textbf{LLM}            & \textbf{Task}    & \textbf{Win Rate (\%)} \\ \midrule
\multirow{2}{*}{\textbf{Llama-3.3-70B}}  & Concrete         & \textbf{56}                     \\
                        & Abstract         & \textbf{52}                     \\ \midrule
\multirow{2}{*}{\textbf{DeepSeek-V3}}  & Concrete         & \textbf{55}                     \\
                        & Abstract         & 49                     \\ \midrule
\multirow{2}{*}{\textbf{ChatGPT-4o}}    & Concrete         & \textbf{51}                     \\
                        & Abstract         & \textbf{53}                    \\ \bottomrule
\end{tabular}
\label{tab:win_rates_LLM}
\vspace{-0.25in}
\end{table}

\begin{table}[t]
\centering
\renewcommand{\arraystretch}{0.7}
\caption{User study for Prompt Enhancement vs. Basic LLMPopcorn prompt.}
\begin{tabular}{@{}llc@{}}

\toprule
\textbf{LLM}            & \textbf{Task}    & \textbf{Win Rate (\%)} \\ \midrule
\multirow{2}{*}{\textbf{User Evaluation}}  & Concrete         & \textbf{66.67}                     \\
                        & Abstract         & \textbf{62.50}                     \\ \midrule
\end{tabular}
\vspace{-0.15in}
\label{tab:user_study}
\end{table}

\noindent\textbf{Implementation details}
\label{sec:implementation}
In this paper, we utilized five state-of-the-art LLMs, including two open-source models, Llama-3.3-70B and Qwen-2.5-72B, and three API-based models, DeepSeek-V3, DeepSeek-R1, and ChatGPT-4o. For efficiency, we apply 4-bit quantization to the open-source LLMs. Additionally, we incorporated three widely-used open-source video generation models: CogVideoX-5B, LTX-Video, and HunyuanVideo. All model weights were obtained from the HuggingFace platform\footnote{\url{https://huggingface.co/}} to ensure consistency and reproducibility. Notably, all open-source LLMs used in this study were instruction-tuned versions. We ensured reproducibility by fixing the random seed for opensourced models, allowing for consistent results. We adopt Faiss\footnote{\url{https://github.com/facebookresearch/faiss}} as a library for matching and indexing. For the prompt enhancement related to RQ2, we search the retrieval-augmented generation (RAG) size from [10, 50, 100]. 
The video generation prompt contains existing video titles and video captions paired with their popularity. All experiments are conducted on one H100 80G GPU. Due to the space limit, the prompt templates and user query datasets used in this paper can be checked from the code repository.

\section{Popular Micro-videos Generation (RQ1)}

\begin{table}[!h]
\centering
\setlength{\tabcolsep}{2pt}
\renewcommand{\arraystretch}{0.7}
\caption{Different LLMs' Win Rate comparisons across concrete and abstract datasets.
Green indicates cases where the Video Generation model outperforms, red signifies losses, and yellow represents ties. BP denotes the basic prompt, WR represents the win rate, and PE refers to prompt enhancement.}
\scalebox{1}{
\begin{tabular}{@{}l cccccc@{}}
\toprule
\multirow{2}{*}{\textbf{LLM}} & \multirow{2}{*}{\textbf{Model Comparison}} 
             & \multicolumn{2}{c}{\textbf{Concrete}} 
             & \multicolumn{2}{c}{\textbf{Abstract}} \\ 
\cmidrule(r){3-4} \cmidrule(r){5-6}
&                           
& \textbf{\,BP\,} 
& \textbf{PE} 
& \textbf{\,BP\,} 
& \textbf{PE} \\ 
\midrule
\multirow{4}{*}{\textbf{Llama-3.3-70B}} 
  & vs. DeepSeek-V3   
     & \cellcolor{red!30}48 & \cellcolor{red!30}48 
     & \cellcolor{red!30}43 & \cellcolor{yellow!30}50 \\

  & vs. ChatGPT-4o  
     & \cellcolor{red!30}49 & \cellcolor{green!30}51 
     & \cellcolor{red!30}49 & \cellcolor{green!30}53 \\

  & vs. DeepSeek-R1    
     & \cellcolor{red!30}46 & \cellcolor{red!30}40 
     & \cellcolor{red!30}45 & \cellcolor{green!30}53 \\

  & vs. Qwen-2.5-72B  
     & \cellcolor{green!30}51 & \cellcolor{green!30}56 
     & \cellcolor{yellow!30}50 & \cellcolor{green!30}55 \\
\midrule

\multirow{4}{*}{\textbf{DeepSeek-V3}} 
  & vs. Llama-3.3-70B 
     & \cellcolor{green!30}52 & \cellcolor{green!30}52 
     & \cellcolor{green!30}57 & \cellcolor{yellow!30}50 \\

  & vs. ChatGPT-4o   
     & \cellcolor{green!30}52 & \cellcolor{green!30}55 
     & \cellcolor{green!30}52 & \cellcolor{green!30}60 \\

  & vs. DeepSeek-R1    
     & \cellcolor{green!30}51 & \cellcolor{red!30}44 
     & \cellcolor{yellow!30}50 & \cellcolor{green!30}58 \\

  & vs. Qwen-2.5-72B  
     & \cellcolor{yellow!30}50 & \cellcolor{green!30}64 
     & \cellcolor{green!30}55 & \cellcolor{red!30}49 \\
\midrule

\multirow{4}{*}{\textbf{ChatGPT-4o}} 
  & vs. Llama-3.3-70B 
     & \cellcolor{green!30}51 & \cellcolor{red!30}49 
     & \cellcolor{green!30}51 & \cellcolor{red!30}46 \\

  & vs. DeepSeek-V3   
     & \cellcolor{red!30}48 & \cellcolor{red!30}45 
     & \cellcolor{red!30}48 & \cellcolor{red!30}40 \\

  & vs. DeepSeek-R1    
     & \cellcolor{red!30}47 & \cellcolor{red!30}43 
     & \cellcolor{red!30}46 & \cellcolor{red!30}48 \\

  & vs. Qwen-2.5-72B  
     & \cellcolor{green!30}52 & \cellcolor{green!30}54 
     & \cellcolor{red!30}49 & \cellcolor{red!30}44 \\
\midrule

\multirow{4}{*}{\textbf{DeepSeek-R1}} 
  & vs. Llama-3.3-70B 
     & \cellcolor{green!30}54 & \cellcolor{green!30}60 
     & \cellcolor{green!30}55 & \cellcolor{red!30}47 \\

  & vs. DeepSeek-V3   
     & \cellcolor{red!30}49 & \cellcolor{green!30}56 
     & \cellcolor{yellow!30}50 & \cellcolor{red!30}42 \\

  & vs. ChatGPT-4o   
     & \cellcolor{green!30}53 & \cellcolor{green!30}57 
     & \cellcolor{green!30}54 & \cellcolor{green!30}52 \\

  & vs. Qwen-2.5-72B  
     & \cellcolor{green!30}56 & \cellcolor{green!30}66 
     & \cellcolor{green!30}55 & \cellcolor{green!30}51 \\
\midrule

\multirow{4}{*}{\textbf{Qwen-2.5-72B}} 
  & vs. Llama-3.3-70B 
     & \cellcolor{red!30}49 & \cellcolor{red!30}44 
     & \cellcolor{yellow!30}50 & \cellcolor{red!30}45 \\

  & vs. DeepSeek-V3   
     & \cellcolor{yellow!30}50 & \cellcolor{red!30}36 
     & \cellcolor{red!30}45 & \cellcolor{green!30}51 \\

  & vs. ChatGPT-4o  
     & \cellcolor{red!30}48 & \cellcolor{red!30}46 
     & \cellcolor{green!30}51 & \cellcolor{green!30}56 \\

  & vs. DeepSeek-R1    
     & \cellcolor{red!30}44 & \cellcolor{red!30}34 
     & \cellcolor{red!30}45 & \cellcolor{red!30}49 \\
\bottomrule
\end{tabular}
}
\label{tab:win_rates_llms_benchmark}
\vspace{-0.15in}
\end{table}

We evaluate three LLMs—Llama-3.3-70B, ChatGPT-4o, and DeepSeek-V3—with CogVideoX as the default video generation model, using concrete and abstract prompt datasets from Section \ref{sec:dataset}. We compare LLMPopcorn-generated video popularity to human-created videos from the Microlens dataset \cite{ni2023content} by matching prompts to semantically similar video titles using an embedding model to reduce thematic bias.
Table \ref{tab:popularity_medians} shows DeepSeek-V3-generated videos achieve median popularity scores of 0.56 (concrete) and 0.73 (abstract), rivaling Microlens’ 0.44 (concrete) and 0.71 (abstract). However, absolute scores remain low, likely due to limitations in open-source video models \cite{liu2024sora}.
Figure \ref{fig:popularity_distribution} reveals DeepSeek-V3’s smoother popularity distribution for abstract prompts, with more videos in mid-to-high ranges, while Llama-3.3-70B and ChatGPT-4o show concentrated low scores with rare outliers. For concrete prompts, model differences are minimal.
Table \ref{tab:comparison_categories} indicates DeepSeek-V3 leads in abstract dataset categories like Anime, Film, and Delicacy, with less dominance in Daily Sharing and Game.
\textbf{(Answer to RQ1)} LLMs enhance micro-video creation by structuring descriptions, with DeepSeek-V3 achieving popularity comparable to or surpassing human-created videos, highlighting AI’s potential for engaging short video generation.

\vspace{-0.1in}

\section{Popularity Enhancement (RQ2)}
\label{sec:PE}
\begin{table}[!h]
\centering
\renewcommand{\arraystretch}{0.7}
\setlength{\tabcolsep}{1pt}
\caption{Different video generation models' Win Rate comparisons across concrete and abstract Datasets.}
\scalebox{1}{
\begin{tabular}{@{}lcccccc@{}}
\toprule
\multirow{2}{*}{\textbf{Video Generators}} & \multirow{2}{*}{\textbf{Comparison}} & \multicolumn{2}{c}{\textbf{Concrete}} & \multicolumn{2}{c}{\textbf{Abstract}} \\ 
\cmidrule(r){3-4} \cmidrule(r){5-6}
&                           & \textbf{\,\,BP\,\,} & \textbf{PE} & \textbf{\,BP\,} & \textbf{PE} \\ \midrule
\multirow{2}{*}{\textbf{CogVideoX-5b}} 
  & vs. LTX-Video   & \cellcolor{red!30}49 & \cellcolor{red!30}38 & \cellcolor{red!30}42 & \cellcolor{red!30}39 \\ 
  & vs. HunyuanVideo & \cellcolor{red!30}44 & \cellcolor{red!30}38 & \cellcolor{red!30}47 & \cellcolor{yellow!30}50 \\ \midrule
\multirow{2}{*}{\textbf{LTX-Video}} 
  & vs. CogVideoX-5b & \cellcolor{green!30}51 & \cellcolor{green!30}62 & \cellcolor{green!30}58 & \cellcolor{green!30}61 \\ 
  & vs. HunyuanVideo   & \cellcolor{red!30}48 & \cellcolor{green!30}52 & \cellcolor{red!30}49 & \cellcolor{green!30}60 \\ \midrule
\multirow{2}{*}{\textbf{HunyuanVideo}} 
  & vs. CogVideoX-5b & \cellcolor{green!30}56 & \cellcolor{green!30}62 & \cellcolor{green!30}53 & \cellcolor{yellow!30}50 \\ 
  & vs. LTX-Video & \cellcolor{green!30}52 & \cellcolor{red!30}48 & \cellcolor{green!30}51 & \cellcolor{red!30}40 \\ \bottomrule
\end{tabular}
}
\label{tab:win_rates_vgs_benchmark}
\vspace{-0.1in}
\end{table}

Table~\ref{tab:win_rates_LLM} compares Prompt Enhancement (PE) to the basic LLMPopcorn prompt using win rates from pairwise comparisons across models. LLama-3.3-70B achieves 56\% (concrete) and 52\% (abstract) win rates, outperforming LLMPopcorn. DeepSeek-V3 scores 55\% (concrete) and 49\% (abstract), while ChatGPT-4o maintains 53\% (abstract), showing PE’s creative compatibility.
A user study with DeepSeek-V3 evaluated PE against LLMPopcorn. Participants blindly compared five video pairs per prompt type, selecting the more popular video. Table~\ref{tab:user_study} shows PE achieved over 60\% win rates for both prompt types.
\textbf{(Answer to RQ2)} PE significantly enhances prompt effectiveness, validated by offline metrics and user studies.

\section{Benchmarking Models (RQ3)}
\label{sec:Bench}
We evaluated five large language models (LLMs): Llama-3.3-70B \cite{touvron2023llama}, Qwen-2.5-72B \cite{yang2024qwen2}, ChatGPT-4o \cite{hurst2024gpt}, DeepSeek-V3, and DeepSeek-R1 \cite{guo2025deepseek}, using their API versions. For video generation, we tested CogVideoX-5B \cite{yang2024cogvideox}, LTX-Video \cite{HaCohen2024LTXVideo}, and HunyuanVideo \cite{kong2024hunyuanvideo}, with instruction-tuned versions for open-source LLMs.

Table~\ref{tab:win_rates_llms_benchmark} shows DeepSeek-V3 and DeepSeek-R1 leading, with DeepSeek-R1 achieving a peak BP WR of 56\% and PE WR of 66\%. Llama-3.3-70B performs strongly in PE WR for abstract tasks (56\%), while Qwen-2.5-72B lags with a PE WR of 34\% against DeepSeek-R1. Table~\ref{tab:win_rates_vgs_benchmark} indicates LTX-Video and HunyuanVideo outperform CogVideoX-5B, both exceeding 50\% win rates in pairwise comparisons, with a peak PE WR of 62\% on the Concrete dataset. \textbf{(Answer to RQ3)} DeepSeek-V3 and DeepSeek-R1 excel among LLMs, consistently leading in Concrete and Abstract scenarios. LTX-Video and HunyuanVideo show comparable, reliable performance in video generation across diverse prompts.

\begin{figure}[!h]
    \centering
   \includegraphics[width=\linewidth]{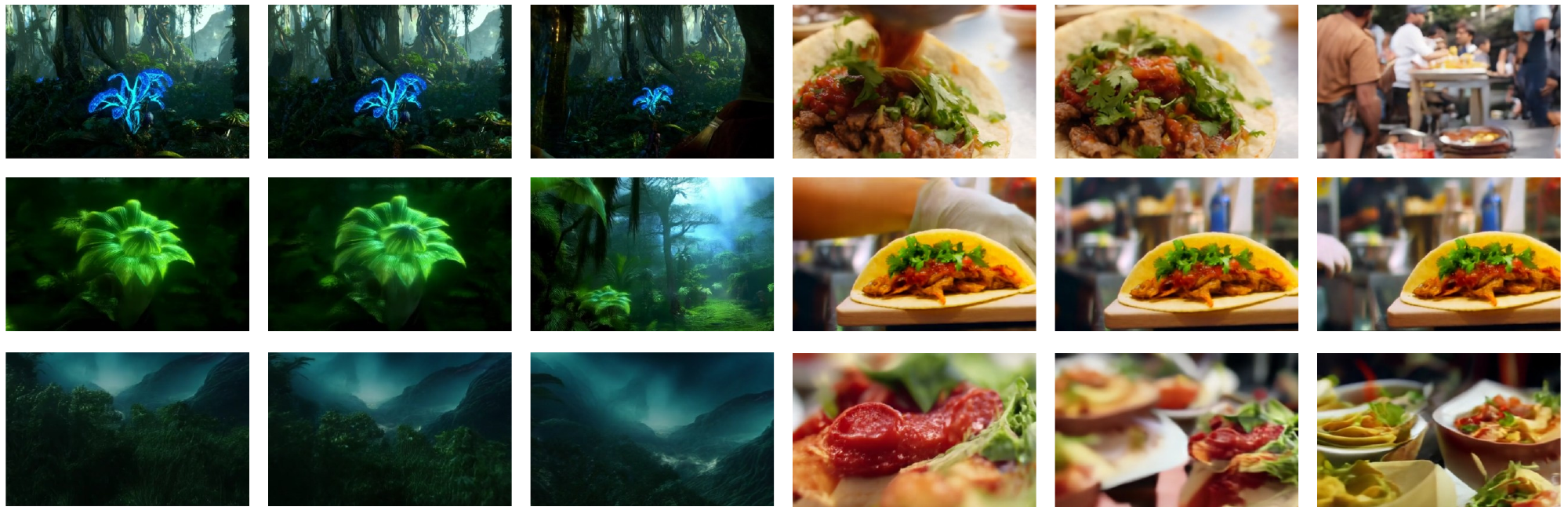}
   \vspace{-0.25in}
    \caption{Video examples from LLMPopcorn, shown from left to right: abstract user queries (1–3) followed by concrete user queries (4–6).}
    \vspace{-0.15in}
    \label{fig:LLMPopcorn_concrete}
\end{figure}

\begin{figure}[!h]
    \centering
   \includegraphics[width=\linewidth]{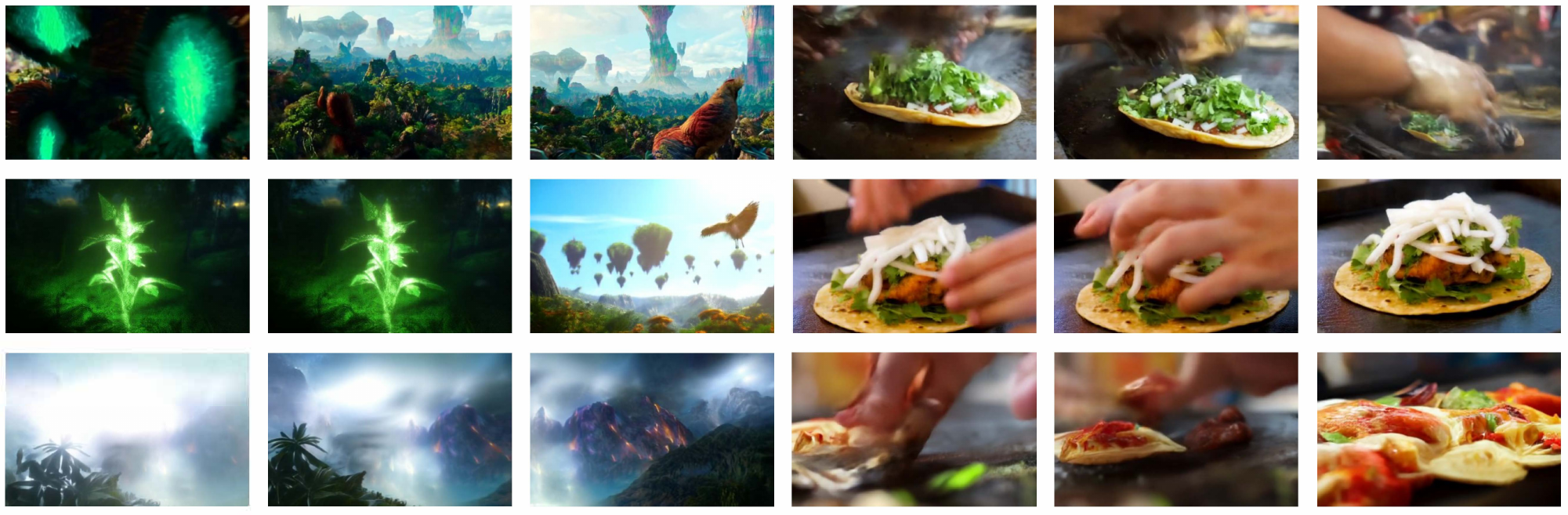}
    \vspace{-0.25in}
    \caption{Video examples from LLMPopcorn with Prompt Enhancement, shown from left to right: abstract user queries (1–3) and concrete user queries (4–6).}
    \vspace{-0.2in}
    \label{fig:LLMPopcorn_abstract}
\end{figure}

\section{Visualization}

To compare different video generation models, we present visualizations generated using two prompts across three models, as shown in Figures~\ref{fig:LLMPopcorn_concrete} and \ref{fig:LLMPopcorn_abstract}.  Each figure includes three rows: videos in the first row are generated by \textbf{CogVideoX-5b}, those in the second row by \textbf{HunyuanVideo}, and those in the third row by \textbf{LTX-Video}. The first prompt is abstract: \textit{``Create a video about the history and evolution of a popular street food."} The second prompt is a concrete prompt: \textit{``Create a video explaining the world-building in Avatar (2009)."} These examples help illustrate how each model handles abstract versus concrete prompts. In all figures, the first three columns represent abstract data, while the last three represent concrete data. From the two figures, Prompt Enhancement may boost popularity by enriching queries to produce dynamic, narrative-driven content. This is shown in its ability to transform abstract prompts into a more diverse range of imaginative scenes (whereas the baseline results are more uniform), while also depicting concrete subjects through engaging actions, like a chef's hands at work, instead of as static objects.

\section{Conclusion}
This empirical study demonstrates that the proposed LLMPopcorn for micro-video generation assisted by state-of-the-art LLMs performs slightly better than ordinary human-created videos, evaluated with an offline popularity predictor.
These results highlight the potential of this approach. To further enhance the LLMPopcorn pipeline, we introduced prompt enhancement (PE), which exhibited strong adaptability across multiple LLMs and video-generation models. Among the tested models, DeepSeek-V3 and DeepSeek-R1 consistently performed best as LLMs, while LTX-Video and HunyuanVideo produced reliable outputs across diverse prompts.  In future work, we plan to explore reinforcement learning and fine-tuning techniques to better align prompts with contextual cues, ultimately improving video quality and engagement. This study pioneers popularity-driven micro-video generation, establishing a strong foundation for future research and innovation.

\footnotesize
\bibliographystyle{IEEEbib}
\bibliography{strings,refs}

\end{document}